\documentclass{article}

\usepackage[nonatbib,preprint]{neurips_2020}

\usepackage[utf8]{inputenc} 
\usepackage[T1]{fontenc}    
\usepackage{hyperref}       
\usepackage{url}            
\usepackage{booktabs}       
\usepackage{amsfonts}       
\usepackage{nicefrac}       
\usepackage{microtype}      

\usepackage{amsmath}
\usepackage{amsthm}
\usepackage{amssymb}
\usepackage{amsfonts}

\usepackage{booktabs} 
\usepackage{subcaption}
\usepackage{url}
\usepackage{listings}
\usepackage{color}
\usepackage{multirow}

\usepackage{caption}
\usepackage{graphicx}

\newtheorem{lemma}{Lemma}

\newtheorem{theorem}{Theorem}
\newtheorem{remark}{Remark}
\newtheorem{corollary}{Corollary}

\usepackage{hyperref}
\usepackage{cleveref}

\crefname{proposition}{Proposition}{Propositions}
\crefname{claim}{Claim}{Claims}

 \newcommand{\squishlist}{
 \begin{list}{$\bullet$}
  { \setlength{\itemsep}{0pt}
     \setlength{\parsep}{1pt}
     \setlength{\topsep}{1pt}
     \setlength{\partopsep}{0pt}
     \setlength{\leftmargin}{1em}
     \setlength{\labelwidth}{1em}
     \setlength{\labelsep}{0.5em} } }
 \newcommand{\squishend}{\end{list}}

\title{Revisiting Weight Initialization of Deep Neural Networks}
\author{Maciej Skorski, Martin Theobald, Alessandro Temperoni}

\begin{document}
\maketitle

\begin{abstract}
The proper initialization of weights is crucial for the effective
training and fast convergence of deep neural networks (DNNs).
Prior work in this area has mostly focused on
{\em balancing the variance
among weights per layer} to maintain stability of (i) the input data propagated forwards through the network and (ii) the loss gradients propagated backwards, respectively.
This prevalent heuristic is however agnostic of dependencies among gradients across the various layers and captures only first-order effects.
In this paper, we propose and discuss an initialization principle that is
based on a {\em rigorous estimation of the global curvature of weights across
layers} by approximating and controlling the norm of their Hessian
matrix. The proposed approach is more systematic and recovers previous
results for DNN activations such as {\em smooth functions}, {\em dropouts}, and {\em ReLU}.
Our experiments on Word2Vec and the MNIST/CIFAR image classification tasks
confirm that tracking the Hessian norm is a useful diagnostic tool which helps to more rigorously initialize weights.
\end{abstract}

\section{Introduction}
\label{sec:intro}
Years of research and practical experience show that parameter initialization is of critical importance for training \emph{neural networks} (NNs), particularly for \emph{deep neural networks} (DNNs), which process their input by stacking several layers of parameterized activation functions.
The main challenge is to determine a good initial ``guess'' of the parameters: small weights may lead to a vanishing effect of (i) the input data being processed forwards, and (ii) the loss gradients being propagated backwards through the network, respectively. Overly large weights, on the other hand, may (i) unduly amplify certain dimensions of the input data in the forward pass, and then in turn (ii) strongly penalize those dimensions during the backward pass. 

As briefly summarized in our related-work discussion, a variety of approaches for weight initialization in DNNs have therefore been explored in the literature~\cite{glorot2010understanding,he2015delving,xu2016revise,hendrycks2016adjusting,hanin2018start,arpit2019initialize,szegedy2013intriguing,virmaux2018lipschitz}. Since both forward and backward propagation of data through a DNN is based on iterative matrix multiplication, current approaches mostly focus on keeping the {\em variance among the weights per layer balanced} (i.e., close to 1), which primarily aims to avoid numerical issues (i.e., vanishing or exploding sums of element-wise matrix multiplications), but they ignore dependencies among weights across the various layers of a DNN.

While the source of difficulty is well-understood, there is no universal remedy: the choice of the initialization scheme is typically studied on a case-by-case basis (depending on the specific architecture and use-case setting) and often under simplifying theoretical assumptions (such as first-order approximations and independence)~\cite{glorot2010understanding,he2015delving,xu2016revise,hendrycks2016adjusting,hanin2018start,arpit2019initialize} .
Even for some relatively simple models like Word2Vec~\cite{mikolov2013efficient,mikolov2013distributed}, we still lack a complete understanding of initialization nuancenses and instead rely on empirically chosen defaults~\cite{kocmi2017exploration}.

\noindent\textbf{Contributions.}
We summarize the contributions of this work as follows.
\squishlist
\item We propose to use {\em second-order methods to estimate the global curvature} (i.e., Hessian) of weights at the initialization time of a DNN.
Our approach thus goes a step further than the existing literature, which relies on local linearization only, and appears more natural from a generic optimization perspective (e.g., in probabilistic inference, the Hessian is widely used for adjusting parameters and diagnosing convergence issues~\cite{salvatier2016probabilistic}).
\item We discuss a framework which can be used to efficiently approximate and control the Hessian norm under our initialization scheme. Under our framework, we derive formulas very close to those proposed before and thereby provide a {\em stronger theoretical justification} for existing initialization schemes such as  {\em smooth activations}~\cite{glorot2010understanding}, {\em dropouts}~\cite{hendrycks2016adjusting}, and {\em Rectified Linear Unit} (ReLU)~\cite{krizhevsky2012imagenet}.
\item Besides our theoretical results, we provide an {\em implementation} of our framework in Tensorflow along with a number of {\em experiments} over both shallow (Word2Vec) and deep (MNIST \& CIFAR) NNs. 
\squishend

\section{Background \& Related Work}
\label{sec:relwork}
Before we review the main weight-initialization schemes proposed in the literature, based on variance flow~\cite{glorot2010understanding,he2015delving,xu2016revise,hendrycks2016adjusting,hanin2018start,arpit2019initialize}
we briefly introduce the key concepts and notation we use through the rest of the paper.

\noindent\textbf{Neural Networks.} From an algebraic perspective, we look at a {\em neural network} (NN) as a chain of mappings of the form
$$z^{(k+1)} = f^{(k)}\left( w^{(k)}\cdot z^{(k)}+b^{(k)}\right)$$ which sequentially processes an {\em input vector} $x=z^{0}$ through a number of {\em layers} $k=0\ldots n-1$. 

We assume that $z^{(k)}$ are real-valued vectors of shape $[d_k]$, {\em weights} $w^{(k)}$ are matrices of shape $[d_{k+1},d_{k}]$, {\em biases} $b^{(k)}$ are of shape $[d_{k+1}]$, and $f^{(k)}$ are (possibly non-linear) {\em activation functions} which are applied element-wisely. The task of learning is to minimize a given {\em loss function} $L(z,t)$ where $z=z^{n}$ is the network output and $t$ is the ground-truth, over the weights $w^{0},\ldots,w^{n-1}$.
Neural networks are optimized with variants of gradient-descent and weights are initialized randomly. Overly small weights make the learning process slow, while too high weights may cause unstable updates and overshooting issues. Good initialization schemes thus aim to find a good balance between the two ends.

\noindent\textbf{Tensor Derivatives.} For two {\em tensors} $y = y_{j_1,\ldots,j_p}$ and $x = x_{i_1,\ldots,i_p}$, of rank $p$ and $q$ respectively, the {\em derivative} 
$D = D_{x} y$ is a tensor of rank $q+p$ with coordinates $D_{j_1,\ldots,j_q,i_1,\ldots,i_p} = \frac{\partial y_{j_1,\ldots,j_q}}{\partial x_{i_1,\ldots,i_p}}$.  
If $y = f(x)$, where $x$ has shape $[n]$ and $y$ has shape $[m]$, then $D_{x}y$ is of shape $[m,n]$ and equals the total derivative of $f$. 

\noindent\textbf{Tensor Products.} Contraction sums over paired indices (axes), thus lowering the rank by 2 (or more when more pairs are specified). For example, contracting positions $a$ and $b$ in $x$ produce the tensor $\sum_{i_a=i_b}x_{i_1 \ldots i_a \ldots i_b\ldots i_p}$ with indices $\{i_1,\ldots,i_p\}\setminus \{i_a,i_b\}$. The dimensions of paired indices should match.
A {\em full tensor product} combines tensors $x$ and $y$ by cross-multiplications $(x\otimes y)_{i_1,\ldots,i_p,j_1,\ldots,j_q} = x_{i_1,\ldots,i_p}\cdot y_{j_1,\ldots,j_p}$, thereby producing a tensor of rank $p+q$. A {\em tensor dot-product} is the full tensor product followed by contraction of two compatible dimensions. For example, the standard matrix product of $A_{i,j}$ and $B_{k,l}$ is the tensor product followed by contraction of $j$ and $k$. We denote the dot-product by $\bullet$, omitting the contracted axes when this is clear from the context.

\noindent\textbf{Chain \& Product Rules.} Tensors obey similar chain and product rules as matrices. Specifically, we have $D_x(A\bullet B) = D_x A \bullet B + A\bullet D_x B$. Also, when $B = f(A(x))$ holds, we have $D_x B = D_{A} f \bullet D_x(A)$. The contraction is over all dimensions of $A$ which match the arguments of $f$.

\noindent\textbf{Spectral Norm.} For any rectangular matrix $A$, the singular eigenvalues are defined as square roots of eigenvalues of $A^T A$ (which is square symmetric, hence positive definite). The {\em spectral norm} then is the biggest singular eigenvalue of $A$.

\subsection{Initialization Based on Variance Flow Analysis}
Glorot and Bengio~\cite{glorot2010understanding} proposed a framework which estimates the variance at different layers in order to maintain the aforementioned balance. The approach assumes that the activation functions approximately behave like the identity function around zero, i.e., $f(u)\approx u$ for small $u$ (this can be easily generalized, see~\cite{he2015delving,xu2016revise}). By linearization, we then obtain:
\begin{align}\label{eq:chain_linearized}
z^{(k+1)}_i\approx  \sum_{j = 1\ldots d_k} w^{(k)}_{i,j}\cdot z^{(k)}_j   + b^{k}_i
\end{align}
In the forward pass, we require $\mathbf{Var}[z^{(k+1)}] \approx \mathbf{Var}[z^{(k)}]$ to maintain the magnitude of inputs until the last layer. In the backward pass, we compute the gradients by recursively applying the chain rule
\begin{align}\label{eq:backward_linearized}
\partial_{z^{(k)}_i} L = \sum_{j=1\ldots d_{k+1}}\partial_{z^{(k+1)}_j} L \cdot \partial_{z^{(k)}_i} z^{(k+1)}_j \approx  \sum_{j=1\ldots d_{k+1}}\partial_{z^{(k+1)}_j} L \cdot w^{(k)}_{j,i}
\end{align}
and want to keep their magnitude, i.e., $\mathbf{Var}\left[\partial_{z^{(k-1)}} L\right] \approx \mathbf{Var}\left[\partial_{z^{(k)}} L\right]$.
Looking at Eq.~\ref{eq:chain_linearized} and \ref{eq:backward_linearized}, we see that the weights $w^{(k)}$ interact with the previous layer during the forward pass and with the following layer during the backward pass. 
The first action is multiplying along the input dimension $d_{k}$, while the second action is multiplying along the output dimension $d_{k+1}$. One can prove that, in general, taking the dot-product with an \emph{independently} centered random matrix along dimension $d$ scales the variance by the factor $d$~\cite{glorot2010understanding,xu2016revise}. Thus, to balance the two actions during the forward and backward pass, one usually chooses the $w^{(k)}$ as i.i.d. samples from a normal distribution $N(\mu, \sigma^2)$ with mean $\mu=0$ and standard deviation
\begin{align}
\sigma[w^{(k)}] = \sqrt{\frac{2}{{d_k}+{d_{k+1}}}}.
\end{align}

Variance-based initialization schemes~\cite{glorot2010understanding,he2015delving,xu2016revise,hendrycks2016adjusting,arpit2019initialize}, however, implicitly assume {\em independence} of weights across layers. To our knowledge, we are first to point out that this is \emph{not true already in first pass}, since back-propagated gradients depend on weights used during the forward pass and also on the input data. As an example, consider a regression setting with two layers and a linear activation function, such that $L = (z-t)^2$, $z = w_2 w_1 x$. Note that $\partial_z L = 2(z-t) = -2(w_2w_1-t)$ here depends on both $w_2$ and $w_1$. To see correlations with the input vector, consider a one-dimensional regression $L=(z-t)^2$, $z=w x$. From Eq. 5 in~\cite{glorot2010understanding}, we should have  $\mathbf{Var}[\partial_w L] = \mathbf{Var}[\partial_{w} z]\cdot \mathbf{Var}[\partial_z L]$ for $w$ with unit variance, but this gives $\mathbf{Var}[2(wx-t)x] = \mathbf{Var}[x]\cdot \mathbf{Var}[2(wx-t)]$. Not only two sides can be a factor away but also the target $t$ can be correlated to the input $x$. In addition to this lack of correlations, this kind of variance analysis also provides only \emph{qualitative} insights, since it does not directly connect the variance estimation to the optimization problem. In fact, we cannot get more quantitative insights, such as estimating the step size, from these first-order methods.

\subsection{Initialization Based on Lipschitzness Estimation}
Recent works~\cite{szegedy2013intriguing,virmaux2018lipschitz} have proposed to estimate Lipschitzness of neural networks in the context of sensitivity analysis. Although not explicitly proposed, in principle such estimates could be adapted to the problem of weight initialization, namely by initializing weights so that the resulting Lipschitz constant is relatively small. Unfortunately, these methods exploit sub-multiplicativity of matrix norms which usually results in overly pessimistic guarantees; for example, for AlexNet (with default initialization), we get an over-estimation by an order of $\sim 10^6$. 


\section{Hessian-Based Weight Initialization}
\label{sec:hessian}
We now present and discuss our suggested weight-initialization scheme by applying a variant of the Hession chain rule across the (hidden) layers $k=0\ldots n-1$ of a neural network, which constitutes the main contribution of our work. 
In general, for training a neural network, variants of gradient-descent are applied in order to update the model parameters $w$ iteratively toward the gradient $g = D_{w} L$ of the loss function. In order to quantify this decrease, we need to consider the {\em second-order approximation}
$$ L(w-\gamma g)\approx L(w) -\gamma\, g^{T}\cdot  g + \frac{\gamma^2}{2}\,g^T\cdot \mathbf{H} \cdot g $$
where $\cdot$ stands for the matrix (or more generally: tensor) dot product. The maximal step size $\gamma^{*}$ guarantees that the decrease equals $\gamma^{*} = \|\mathbf{H}\|^{-1}$~\cite{Goodfellow-et-al-2016} where $\|\mathbf{H}\|$ is the Hessian norm, i.e., its maximal eigenvalue. In other words, if we want to train with a constant step size, then we need to control the Hessian. We therefore propose the following paradigm:
\begin{quote}
\emph{Good weight initialization controls the Hessian:} we initialize the weights $w$ such that $\|\mathbf{H}_{w^{(k)}}\| \approx 1$.
\end{quote}
Moreover, we only make the following mild assumption about the loss functions:
\begin{quote}
\emph{Admissible loss functions:} the loss function must satisfy $f(0)=0$ and $f''(0)=0$. Note that this is the case for all standard functions: linear, sigmoid, tanh, relu.
\end{quote}
Finally, our techniques aim to approximate the global curvature of weights up to leading terms. These approximations are accurate under the following mild assumption:
\begin{quote}
\emph{Relatively small inputs:} we have $\|z^{(k)}\| \leqslant c$ for all layers $k$, for some small constant $c$ (e.g., $c=0.1$).
\end{quote}
Note that the latter simply ensures stability of the forward pass and is implicitly assumed so also in the variance flow analysis (which however assumes a linear regime).


Before presenting our results, we need to introduce some more notation. 
Let $F^{(k)} = z^{(k)}$ be the input of the $k$-th layer.
Let $A^{(k)} = D_{u^{(k)}}z^{(k+1)}$ be the derivative of the forward activation at the $k$-th layer, with respect to the output before activation $u^{(k)} = w^{(k)}\cdot z^{(k)}+b^{(k)}$. Let $B^{k+1} = D_{z^{(k+1)}} z^{(n)}$ be the output derivative back-propagated to the input of  the $(k+1)$-th layer. 
Let $\mathbf{H}_z = D^2_{z} L(z,t)$ be the loss Hessian with respect to the predicted value $z$. Finally, let $\mathbf{H}_{w} = D^2_{w}L(z^{(n)},t)$ be the loss Hessian with respect to the weights $w$.

\subsection{Approximation via Hessian Chain Rule}
The Hessian of the loss function over its domain is usually very simple and has nice properties. This however changes when a neural network reparameterizes the problem by a complicated dependency of the output $z$ on the weights $w$. We thus have to answer the following question: how does the dependency of the network output on the weights affect the curvature? 

In general, if $z=z(w)$ is a reparameterization, then it holds that
\begin{align}
\underbrace{D^2_{w} L(z(w),t)}_{\text{reparameterized Hessian}} = D^{2}_{z} L(z,t) \underbrace{\bullet D_{w}z(w)\bullet D_{w}z(w)}_{\text{linearization effect}} +  D_z L(z,t)\underbrace{\bullet D_{w}^2 z(w)}_{\text{curvature effect}}
\end{align}
where bullets denote tensor dot-products along the appropriate dimensions. This is more subtle than back-propagation of first derivatives, because both first- and second-order effects have to be captured. The main contribution of this work thus is the following result, which in its essence states that, usually, the curvature effect contributes less than the linearization effect.
\begin{theorem}[Approximated hessian chain rule for neural networks]\label{thm:hessian_chainrule}
With notation as above, the loss Hessian $\mathbf{H}_{w^{(k)}}$ with respect to the weights $w^{(k)}$ satisfies (up to the leading term)
\begin{align}
 \mathbf{H}_{w^{(k)}}[g,g] \approx v^T \cdot \mathbf{H}_z \cdot v ,\quad v = B^{(k)}\cdot A^{(k)}\cdot g \cdot F^{(k)}
\end{align}
where products are standard matrix products. More precisely, the approximation holds up to a \emph{third-order} error term $\sim f'''c^3\cdot\|g\|^2$ where $f'''$ is the bound on the third derivative of the activation functions and $c$ is the bound on the inputs $x$. The leading term then is of order $\sim c^2\cdot\|g\|^2$.
\end{theorem}

We observe the following important properties. 

\begin{remark}[Low Computational Complexity]
Computing the hessian approximation is of cost comparable to backpropagation. The only hessian we need is the loss/output hessian which is usually small ($K^2$ for classification of $K$ classes). 
\end{remark}

\begin{remark}[Beyond MLP Model]
We formulated the result for densely-connected networks but the approximation holds in general with $v = D_{w^{(k)}} z^{(n)}\bullet g$ (as we will see in empirical evaluation).
\end{remark}
\begin{remark}[Perfect approximation for ReLU networks]
We have exact equality for activations with $f'' = 0$ such as variants of ReLU (see Section~\ref{sec:relu}). 
\end{remark}
\begin{remark}[Good approximation up to leading terms]
Regardless of the activation function, the error term is of smaller order under our assumption of relatively small inputs. 
\end{remark}
We provide an empirical validation of \Cref{thm:hessian_chainrule} in our experiments in \Cref{sec:experiments}.


\subsection{Approximation via Jacobian Products}
From the previous subsection, we are left with the linearization effect of the chain rule, which can be further factored. This reduces the problem to controlling the \emph{products of the hidden layers' Jacobians}. 
\begin{theorem}[Hessian factorized into Jacobians]\label{eq:jacobian_products}
Up to third-order terms in $z^{(i)}$, we can factorize $v$ from \Cref{thm:hessian_chainrule} into
\begin{align}\label{eq:full_factorization}
v \approx \mathbf{J}^{(n-1)}\cdot \ldots \mathbf{J}^{(k+1)}\cdot A \cdot g\cdot \mathbf{J}^{(k-1)}\cdot \ldots \mathbf{J}^{(0)}\cdot z^{(0)}
\end{align}
where $\mathbf{J}^{k} = D_{z^{(k)}}z^{(k+1)}$ is the derivative of the output with respect to the input at the $k$-th layer. In particular, the Hessian's dominant eigenvalue scales by a factor of at most $\|v\|^2$ where
\begin{align}
\|v\| \leqslant \|\underbrace{\mathbf{J}^{(n-1)}\cdot \ldots \cdot\mathbf{J}^{(k+1)}}_{\text{backward product}}\| \cdot \|A\|\cdot  \|\underbrace{\mathbf{J}^{(k-1)}\cdot \ldots\cdot \mathbf{J}^{(0)} \|}_{\text{forward product}} \cdot \|z^{(0)}\|.
\end{align}
\end{theorem}
The norm of the matrix product $\|J_k\ldots J_1\|$ is computed as the maximum of the vector norm $\|J_k\cdots J_1 \cdot v\|$ over vectors $v$ with unit norm.  Given this result, a good weight intialization thus aims to make the backward and forward products having a norm close to one.
\begin{remark}[Connection to products of random matrices]
Note that our problem closely resembles the problem of \emph{random matrix products}~\cite{kargin2010products}. This is because Jacobians for smooth activation functions are simply random-weight matrices.
\end{remark}

\begin{remark}[Connection to spectral norms]
Further, it is possible to estimate the product of random matrices by the product of their spectral norms. In particular, the spectral norm of a random $m\times n$ matrix with zero-mean and unit-variance entries is $\frac{1}{\sqrt{m}+\sqrt{n}}$ on average~\cite{silverstein1994spectral}. For the Gaussian case, this can be found precisely by Wishart matrices~\cite{edelman1988eigenvalues}. This however is overly pessimistic for long products.
\end{remark}


\section{Relationship to Existing Initialization Schemes}
\label{sec:initialization}

We next discuss the relationship of our Hessian-based weight initialization scheme to a number of previous schemes, namely {\em smooth 
activations}~\cite{glorot2010understanding}, {\em dropout}~\cite{hendrycks2016adjusting}, and {\em ReLU}~\cite{krizhevsky2012imagenet}.

\subsection{Smooth Activations}
We first formulate the following lemma.
\begin{lemma}[Dot-product by random matrices]
Let $w$ be a random matrix of shape $[n,m]$, with zero-mean entries and a variance of $\sigma^2$. Let $z$, $z'$ be independent vectors of shape $[m]$ and $[n]$, respectively. It then holds that:
\begin{align}
\mathbf{E} \| w  \cdot z\|^2 &= n\sigma^2 \cdot \mathbf{E}\|z\|^2 \\
\mathbf{E} \| z' \cdot w\|^2 &= m\sigma^2 \cdot \mathbf{E}\|z'\|^2
\end{align}
\end{lemma}
Using this, we can estimate the growth of Jacobian products in \Cref{eq:jacobian_products} as follows.

\begin{corollary}[Smooth activations~\cite{glorot2010understanding}]
Consider activation functions such that $f'(0)=1$.  Then $\mathbf{J}^{(k)} \approx w^{(k)}$ (up to leading terms) and
the norm of the forward product is stable when
\begin{align}
\mathbf{Var}[w^{(k)}] = \frac{1}{d_{k+1}},
\end{align}
while the norm of the backward product is stable when
\begin{align}
\mathbf{Var}[w^{(k)}] = \frac{1}{d_{k}}.
\end{align}
As a compromise, we can choose $\mathbf{Var}[w^{(k)}]  = \frac{2}{d_{k+1}+d_{k}}$.
\end{corollary}
Note that we exploit the fact that (up to leading terms) Jacobians of smooth activation functions are independent from any other components.

\subsection{Dropouts}
Dropouts (i.e., inactive neurons) can be described by a \emph{randomized function} $f_p$ which, for a certain dropout probability $p$, multiplies 
the input by $B_{1-p}\cdot \frac{1}{1-p}$ where $B_{1-p}$ is a Bernoulli random variable with parameter $1-p$. 
The Jacobian then is precisely given by:
\begin{align}
\mathbf{J} = w = \mathsf{diag}(B_1,\ldots,B_d),\quad B_i \sim\mathsf{Bern}(1-p)
\end{align}
When multiplying from left or right, this scales the norm square by $(1-p)^{-2}\cdot \mathbf{E}[\mathsf{Bern}(1-p)]^2 = 1-p$. Thus, we obtain the following corollary.
\begin{corollary}[Initialization for dropout]
Let $1-p$ be the \emph{keep rate} of a dropout. Let $\sigma^2$ be the initialization variance without dropouts, then it should be corrected as:
\begin{align}
\sigma' = \sigma / \sqrt{1-p}
\end{align}
\end{corollary}
This corresponds to the analysis in~\cite{hendrycks2016adjusting}, except that they suggest a different correction factor for back-propagation. 

\subsection{ReLU}
\label{sec:relu}
Rectified Linear Unit (ReLU)~\cite{krizhevsky2012imagenet} is a non-linear activation function given by $f(u) = \mathrm{max}(u,0)$.
Consider a layer such that $z' = f(u)$, $u=w\cdot z$ where $w$ is zero-centered with a variance of $\sigma^2$ and again of shape $[n, m]$, while $z$ is of shape $m$. We then have $\mathbf{J} = D_{z} z' = \mathsf{diag}(f'(u))\cdot w$.

For the forward product, we consider $\mathbf{J}\cdot z = \mathsf{diag}(f'(u))\cdot u$ . This scales the norm of $u$ by $\frac{1}{2}$ when $u$ is symmetric and zero-centered, which is true when also $w$ is symmetric and zero-centered. The norm of $z$ is thus changed by $\frac{n\sigma^2}{2}$.

For the backward product, on the other hand, we have to consider $v\cdot J\cdot  \mathbf{J}$, where $J$ is the Jacobian product for the subsequent layers and possibly depends on $u$. However, if the next layer is initialized with i.i.d. samples, the output distribution only depends on the number of active neurons $r = \#\{i: u_i = 1\}$. Conditioned on this information, the following layers are independent from $\mathbf{J}$. Given $r$, the squared backward product norm thus changes by the factor $r/n \cdot m\sigma^2$. Since $\mathbf{E}[r] = n/2$, the scaling factor is $\frac{m\sigma^2}{2}$.

\begin{corollary}[ReLU intialization~\cite{he2015delving}]
The initialization variance $\sigma^2$ in the presence of ReLU should thus be corrected as:
\begin{align}
\sigma' = \frac{\sigma}{\sqrt{2}}
\end{align}
\end{corollary}

We remark that similar techniques can also be used to derive formulas for weighted ReLU~\cite{arpit2019initialize}.

\section{Experiments}
\label{sec:experiments}
We conducted the following experiments to confirm the theoretical findings stated in the previous two sections. To cover a broad and diverse range of experiments, we trained our models on the MNIST and CIFAR image datasets available in Keras, as well as on a large collection of Wikipedia sentences for training a Word2Vec model. 

\noindent\textbf{Implementation.} All models were coded in Python using the \texttt{Tensorflow 1.15}~\cite{tensorflow2015-whitepaper} and \texttt{Numpy 1.8} core libraries. Random-number generators of all libraries were properly seeded to ensure reproducibility.\footnote{Notebooks for all experiments run on Google Colab and are available from the supplementary material.} 

\noindent\textbf{Hessian Calculation.}
Hessian calculations are not well supported by the Tensorflow API, even in its most recent release \texttt{2.2.0}.
The default implementation under \texttt{tf.hessians} does not work with fused operations, including certain loss functions such as the sparse cross-entropy used in classification~\cite{tf_bug_hess_fusedops}; moreover it 
doesn't support evaluating hessian products without explicitly creating the whole hessian which quickly leads to out-of-memory issues;
batch mode is also not supported.
The parallel computation of components is supported for jacobians only as of very recently~\cite{agarwal2019static}, but also does not work when composing higher-order derivatives~\cite{tf_bug_loops_derivatives}. 
When implementing our approximation we thus resort to a hybrid solution by expressing Hessians as a composition of sequential gradients which are followed by a parallelized computation of the Jacobians. 

\subsection{Correlations among Loss and Layer Gradients}
As previously argued, we expect the gradient of the loss with respect to the output $\frac{\partial L}{\partial z^{(n)}}$ and the layer-to-layer gradients $\frac{\partial z^{(i)}}{z^{(i-1)}}$ to be correlated. We therefore prepared an experiment to demonstrate these correlations based on a simple 3-layer NN with Glorot's initialization scheme over the MNIST dataset.
We re-ran the initialization a large number of times and used \emph{Pearson's r correlation test} as it is implemented in $\texttt{SciPy}$~\cite{2020SciPy-NMeth} to estimate the dependencies, each using different random seeds. Our findings confirm (i) {\em very significant correlations} among components of the loss gradient and the network output, as well as (ii) {\em significant correlations} between the loss/output gradient and the output/layer gradients. The experiment is summarized in \Cref{tab:experiments}.
\begin{table}[ht!]
\centering
\resizebox{\columnwidth}{!}{%
{\small
\begin{tabular}{|c|c|c|c|c|c|}
\toprule
Number of seeds & Samples size & Tested Gradients & Dependency Detected & Comment\\
\toprule
\multirow{2}{*}{10} & $10^4$ & Loss/Output (diff. components)  & 10/10 times & avg. $p$-value $\sim 10^{-5}$ (strong evidence) \\
& $10^5$ & Loss/Output6 vs. Output6/Output5 & 9/10 times & avg. $p$-value $\sim 2\cdot 10^{-2}$  (strong--weak evidence) \\
\bottomrule
\end{tabular}}
}
\caption{Correlations among the components of the back-propagation equation for a simple 3-layer NN. Dependencies tested with \emph{Pearson's r correlation test} at 95\% significance.}
\label{tab:experiments}
\end{table}
\vspace*{-5pt}

\subsection{Error in Hessian Approximation}
In \Cref{thm:hessian_chainrule}, we claim that, up to leading orders of magnitude, the curvature effect can be neglected. We verified this empirically by fixing a dense NN and comparing the true and approximated Hessian value at its initialization. The two Hessians are compared by evaluating their quadratic forms on a randomly chosen direction. Initialization is then restarted several times to estimate the fraction of cases when the approximation is within a required relative error.

We empirically find that the approximation is of good quality, in a sense that it estimates Hessians within the correct order of magnitude~\footnote{This is sufficient in the context of weight initialization. For example, existing initialization schemes cannot guarantee accurate constants in their estimates due to architecture heterogeneity and non-linear effects. }. This has been confirmed on different architectures and activation functions, using both stochastic gradient-descent (SGD) as well as the mini-batch variant with the learning rate of $0.01$. \Cref{tab:hess_approx_minib_table} and \Cref{tab:hess_approx_sgd_table} depict the percentage of samples for which the approximation falls within the tolerance thresholds ($rtol$) over a dense 5-layer NN on the MNIST dataset, and a CNN net (a modification of the LeNet network~\cite{lecun1998gradient}) on the CIFAT dataset. Glorot's initialization scheme and the mini-batch SGD were used for training.

\noindent
\begin{minipage}[b]{.48\textwidth}
\centering
{\small
\begin{tabular}{|c|c|c|c|}
\toprule
Layer & $rtol \leqslant 0.5$ & $rtol\leqslant 1$ & $rtol \leqslant 1.5$  \\
\midrule
Dense 1 & 41\% & 78\% & 92\% \\
Dense 2 & 79\% & 97\% & 97\% \\
Dense 3 & 81\% & 100\% & 100\% \\
Dense 4 & 100\% & 100\% & 100\%\\
Dense 5 & 100\% & 100\% & 100\% \\
\bottomrule
\end{tabular}}
\captionof{table}{Approximation quality of \Cref{thm:hessian_chainrule} for a dense network with $\mathrm{tanh}$ as activation functions on MNIST, with mini-batches of size 32.}
\label{tab:hess_approx_minib_table}
\end{minipage}
\hspace{4mm}
\begin{minipage}[b]{.48\textwidth}
\centering
{\small
\begin{tabular}{|c|c|c|c|}
\toprule
Layer & $rtol \leqslant 1$ & $rtol\leqslant 2$ & $rtol \leqslant 5$  \\
\midrule
Convolution1 & 65\% & 79\% & 92\% \\
Convolution2 & 73\% & 86\% & 95\% \\
\bottomrule
\end{tabular}}
\captionof{table}{Approximation quality of \Cref{thm:hessian_chainrule} for a CNN network with $\mathrm{tanh}$ as activation functions on CIFAR, with mini-batches of size 32.}
\label{tab:hess_approx_sgd_table}
\end{minipage}

\subsection{Hessian Initialization and Convergence}

\noindent\textbf{MNIST.} This model is trained on the MNIST~\cite{mnisthandwrittendigit} dataset and has 2 hidden layers using ReLU activation functions to process 28x28-point images and predict their labels. The loss function is categorical cross-entropy.


\begin{minipage}{\textwidth}
\centering
\begin{minipage}[b]{0.45\textwidth}
\includegraphics[width=0.8\textwidth]{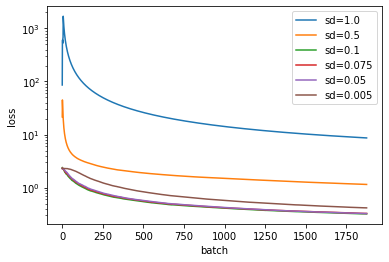}
\captionof{figure}{Loss during training (avg. per batch), depending on weight initialization.}
\label{fig:mnist_hess_init_summary}
\end{minipage}
\quad
\begin{minipage}[b]{0.45\textwidth}
\resizebox{\textwidth}{!}{
\begin{tabular}{|c|c|c|c|}
\toprule 
Weights Std. &Hessian 1 & Hessian 2 & Training Loss \\
\midrule
1.000 & 24.48 & 10.06 & 8.65 \\
0.500 & 12.88 & 10.09 & 1.15 \\
0.100 & 2.34 & 2.76 &  0.32 \\
0.075& 1.18 & 1.49 & 0.33 \\
0.050& 0.49 & 0.65 & 0.33 \\
0.005 & 0.005 & 0.006 & 0.42 \\
\bottomrule
\end{tabular}
}
\vspace{3mm}
\captionof{table}{Loss Hessians (biggest eigenvalues) wrt. hidden layer weights at initialization vs. training loss after two epochs, depending on weight initialization.}
\label{tab:mnist_hess_init_summary}
\end{minipage}
\vspace{0.1cm}
\end{minipage}
As shown in \Cref{fig:mnist_hess_init_summary} and \Cref{tab:mnist_hess_init_summary}, the best standard deviations ($Std.$) correspond to Hessian norms close to 1. We can also see that for larger values of standard deviations, training may not converge; while for smaller values, it converges much slower.

\noindent\textbf{Convolutional Neural Network.} This model is based on a modified LeNet network and has two convolutional layers. We trained it on the CIFAR10~ \cite{cifar} dataset which consists of 50,000 labeled images of 32x3x3-point resolution. 
We approximated the loss Hessians with respect to the convolutional kernels at the initialization, thereby restarting initialization and testing different seeds. Similarly as before, we reach the conclusion that hessians 

\begin{minipage}{\textwidth}
\centering
\begin{minipage}[b]{0.45\textwidth}
\includegraphics[width=0.8\textwidth]{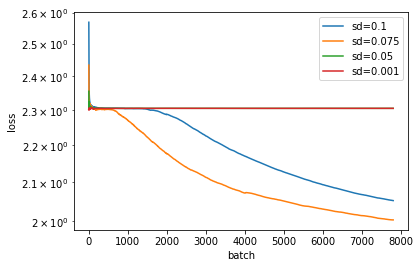}
\captionof{figure}{Loss during training (avg. per batch), depending on weight initialization.}
\label{fig:mnist_hess_init_summary}
\end{minipage}
\quad
\begin{minipage}[b]{0.45\textwidth}
\resizebox{\textwidth}{!}{
\begin{tabular}{|c|c|c|c|}
\toprule 
Weights Std. &Hessian 1 & Hessian 
2 & Training Loss \\
\midrule
0.100 & 1.47 & 1.43 &  1.91 \\
0.075& 0.81 & 0.79 & 1.92 \\
0.050& 0.36 & 0.35 & 2.30 \\
0.001 & 0.0001 & 0.0001 & 2.30 \\
\bottomrule
\end{tabular}
}
\vspace{3mm}
\captionof{table}{Loss Hessians (biggest eigenvalues) wrt. hidden layer weights at initialization vs. training loss after two epochs, depending on weight initialization.}
\label{tab:mnist_hess_init_summary}
\end{minipage}
\vspace{0.1cm}
\end{minipage}


\subsection{Word2Vec Shallow NN with Wikipedia Sentences}
Finally, we also observed the similar relation between the Hessian initialization and convergence for a Word2Vec-based skip-gram model~\cite{mikolov2013efficient}, which can be seen as a shallow NN with only one hidden layer.  As training data, we used 7M of clauses (short sentences) extracted from a recent dump of English Wikipedia articles which were pre-processed with a pipeline of NLP and information-extraction tools using AIDA~\cite{DBLP:conf/www/NguyenHTW14} for named-entity recognition and ClausIE~\cite{DBLP:conf/www/CorroG13} for clause detection. 

\section{Conclusions}

We have discussed how to approximate Hessians of loss functions for neural network, and how to use them to get insights into weight initialization. The main theoretical finding is that this approximation explains initialization schemes developed previously based on a heuristic approaches. Besides theoretical results we provide an empirical validation of these ideas.

\newpage

\bibliographystyle{plain}
\bibliography{citations}

\end{document}